\documentclass[10pt,twocolumn,letterpaper]{article}

\usepackage{wacv}              
\usepackage{pifont}
\newcommand{\cmark}{\ding{51}}%
\newcommand{\xmark}{\ding{55}}%
\usepackage{caption}

\usepackage[accsupp]{axessibility}
\usepackage{graphicx}
\usepackage{amsmath}
\usepackage{amssymb}
\usepackage{booktabs}
\usepackage{multirow}
\usepackage{adjustbox}
\usepackage{pgfplots}
\pgfplotsset{compat=1.17}
\usepackage{tikz}
\usepackage{pgfplots}
\DeclareUnicodeCharacter{2212}{−}
\usepgfplotslibrary{groupplots,dateplot}
\usetikzlibrary{patterns,shapes.arrows}
\usepackage[ruled,vlined]{algorithm2e}

\usepackage[pagebackref,breaklinks,colorlinks]{hyperref}

\usepackage[capitalize]{cleveref}
\crefname{section}{Sec.}{Secs.}
\Crefname{section}{Section}{Sections}
\Crefname{table}{Table}{Tables}
\crefname{table}{Tab.}{Tabs.}

\def\wacvPaperID{1966} 
\def\confName{WACV}
\def\confYear{2024}

\begin{document}
\definecolor{color1_boundary1}{HTML}{616BBD}
\definecolor{color1_boundary2}{HTML}{AD0000}
 \title{Robust Feature Learning and Global Variance-Driven Classifier Alignment for Long-Tail Class Incremental Learning}

\author{Jayateja Kalla and Soma Biswas\\
Department of Electrical Engineering\\
Indian Institute of Science, Bangalore, India.\\
{\tt\small \{jayatejak, somabiswas\}@iisc.ac.in}
}
\maketitle

\begin{abstract}
This paper introduces a two-stage framework designed to enhance long-tail class incremental learning, enabling the model to progressively learn new classes, while mitigating catastrophic forgetting in the context of long-tailed data distributions. Addressing the challenge posed by the under-representation of tail classes in long-tail class incremental learning, our approach achieves classifier alignment by leveraging global variance as an informative measure and class prototypes in the second stage. This process effectively captures class properties and eliminates the need for data balancing or additional layer tuning. Alongside traditional class incremental learning losses in the first stage, the proposed approach incorporates mixup classes to learn robust feature representations, ensuring smoother boundaries. The proposed framework can seamlessly integrate as a module with any class incremental learning method to effectively handle long-tail class incremental learning scenarios. Extensive experimentation on the CIFAR-100 and ImageNet-Subset datasets validates the approach's efficacy, showcasing its superiority over state-of-the-art techniques across various long-tail CIL settings. 
Code is available at \url{https://github.com/JAYATEJAK/GVAlign}.
\end{abstract}
\vspace{-1em}
\section{Introduction}
\label{sec:intro}

\begin{figure}[t!]
\captionsetup{aboveskip=1pt, belowskip=-3pt}
\centering
\begin{tikzpicture}
\definecolor{color0}{HTML}{E8DED2}
\definecolor{color1}{HTML}{A3D2CA}
\definecolor{color2}{HTML}{5EAAA8}
\definecolor{color3}{HTML}{056676}
\definecolor{color4}{HTML}{F6635C}
\begin{groupplot}[group style={group size=1 by 2,  vertical sep=10ex}, title style={anchor=north, yshift=2ex}, width=10cm]
\nextgroupplot[
axis line style={white!80!black},
tick align=outside,
title={a). Shuffled Long-Tail},
x grid style={white!80!black},
xmajorgrids,
xmajorticks=true,
xmin=-0.48, xmax=2.48,
xtick style={color=white!15!black},
xtick={0,1,2},
xticklabels={Long Classes,Tail Classes,All Classes},
xticklabel style={
        anchor=north, 
        yshift=0pt, 
        align=center,
    },
x label style={at={(axis description cs:0.1,1.4)},anchor=north},
xtick pos=left,
y grid style={white!80!black},
y label style={at={(axis description cs:-0.10,.5)}},
ylabel={Accuracy (\%)},
ymajorgrids,
ymajorticks=true,
ymin=05, ymax=70,
ytick style={color=white!15!black},
ytick={15,30,45,60,75},
ytick pos=left,
width = 8cm, height = 4.8cm,
legend style={at={(0.45,+1.2)},  font=\small,legend cell align={left},/tikz/every even column/.append style={column sep=0.1cm},
anchor=south,legend columns=3},
]
\draw[draw=white,fill=color0,opacity=1] (axis cs:-0.3,0) rectangle (axis cs:-0.1,63.8);
\addlegendimage{line width=1.5mm,color=color0}
\addlegendentry{CE}
\draw[draw=white,fill=color0,opacity=1] (axis cs:0.7,0) rectangle (axis cs:0.9,16.12);
\draw[draw=white,fill=color0,opacity=1] (axis cs:1.7,0) rectangle (axis cs:1.9,39.96);
\draw[draw=white,fill=color1,opacity=1] (axis cs:-0.1,0) rectangle (axis cs:0.1,60);
\addlegendimage{line width=1.5mm,color=color1}
\addlegendentry{CE + LWS (ECCV'22)}
\draw[draw=white,fill=color1,opacity=1] (axis cs:0.9,0) rectangle (axis cs:1.1,26);
\draw[draw=white,fill=color1,opacity=1] (axis cs:1.9,0) rectangle (axis cs:2.1,43);
\draw[draw=white,fill=color3,opacity=1] (axis cs:0.1,0) rectangle (axis cs:0.3,68.28);
\addlegendimage{line width=1.5mm,color=color3}
\addlegendentry{CE + GVAlign (Ours)}
\draw[draw=white,fill=color3,opacity=1] (axis cs:1.1,0) rectangle (axis cs:1.3,26.04);
\draw[draw=white,fill=color3,opacity=1] (axis cs:2.1,0) rectangle (axis cs:2.3,47.2);

\nextgroupplot[
axis line style={white!80!black},
tick align=outside,
title={b). Ordered Long-Tail},
x grid style={white!80!black},
xmajorgrids,
xmajorticks=true,
xmin=-0.48, xmax=2.48,
xtick style={color=white!15!black},
xtick={0,1,2},
xticklabels={Long Classes,Tail Classes,All Classes},
xticklabel style={
        anchor=north, 
        yshift=0pt, 
        align=center,
    },
x label style={at={(axis description cs:1.5,0.4)},anchor=north},
xtick pos=left,
y grid style={white!80!black},
y label style={at={(axis description cs:-0.10,.5)}},
ylabel={Accuracy (\%)},
ymajorgrids,
ymajorticks=true,
ymin=40, ymax=80,
ytick style={color=white!15!black},
ytick={15,30,45,60,75},
ytick pos=left,
width = 8cm, height = 4.8cm,
legend style={at={(0.45,+1.2)},  font=\small,legend cell align={left},/tikz/every even column/.append style={column sep=0.1cm},
anchor=south,legend columns=2},
]
\draw[draw=white,fill=color0,opacity=1] (axis cs:-0.3,0) rectangle (axis cs:-0.1,71.4);
\draw[draw=white,fill=color0,opacity=1] (axis cs:0.7,0) rectangle (axis cs:0.9,46.68);
\draw[draw=white,fill=color0,opacity=1] (axis cs:1.7,0) rectangle (axis cs:1.9,59.1);
\draw[draw=white,fill=color1,opacity=1] (axis cs:-0.1,0) rectangle (axis cs:0.1,68);
\draw[draw=white,fill=color1,opacity=1] (axis cs:0.9,0) rectangle (axis cs:1.1,51.7);
\draw[draw=white,fill=color1,opacity=1] (axis cs:1.9,0) rectangle (axis cs:2.1,59.8);
\draw[draw=white,fill=color3,opacity=1] (axis cs:0.1,0) rectangle (axis cs:0.3,77.4);
\draw[draw=white,fill=color3,opacity=1] (axis cs:1.1,0) rectangle (axis cs:1.3,53.4);
\draw[draw=white,fill=color3,opacity=1] (axis cs:2.1,0) rectangle (axis cs:2.3,65.4);


\end{groupplot}

\end{tikzpicture}

\caption{The performance of long, tail, and all classes is illustrated for two long-tail distributions as proposed in \cite{longtail_cil}. It is evident that training the Learnable Weight Scaling (LWS) layer with cross-entropy (CE) loss leads to a reduction in performance for the long classes, while simultaneously improving the performance of the tail classes. In contrast, our proposed approach, which leverages robust features and classifier alignment, exhibits an enhancement in the performance of both long and tail classes, thereby improving the overall all classes performance.}
\label{fig_motivation1}
\vspace{-1em}
\end{figure}
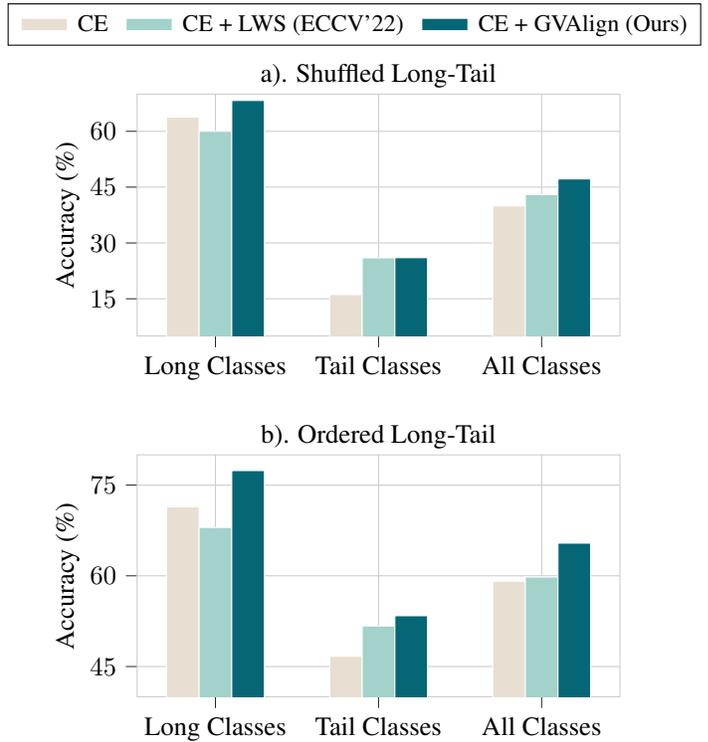
\begin{figure*}[t]
\hspace*{-0.8cm} 
    \centering
     
        \includegraphics[width=\linewidth]{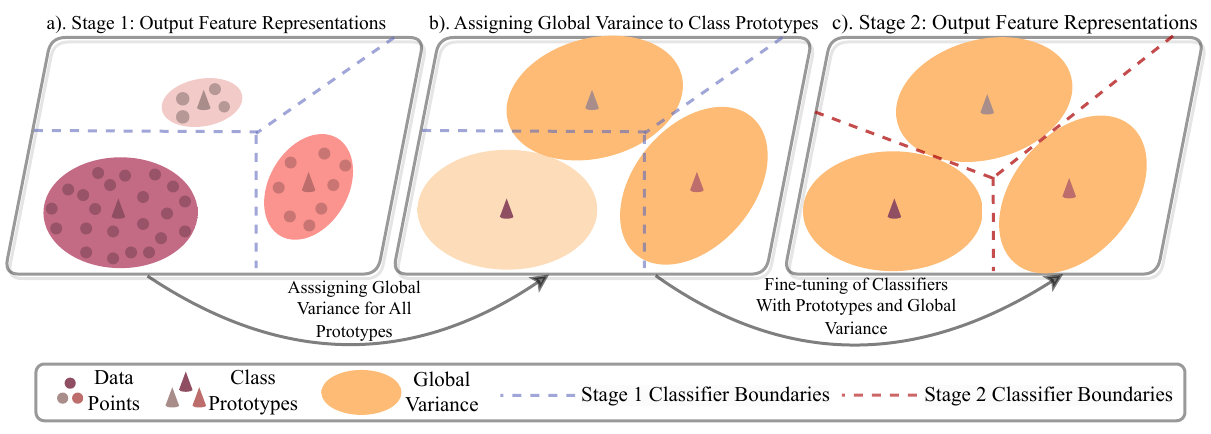}
        \vspace{-1em}
        
        \caption{The leftmost figure `a' illustrates the robust feature representations obtained in stage 1. Once robust representations are acquired, the global variance is assigned across all prototypes figure `b'. Sample feature representations drawn using prototypes and these global variance are then utilized to align the classifiers in stage 2. The rightmost figure `c' depicts the aligned classifiers achieved after stage 2. The global variance signifies the data covariance of the base class with the highest number of samples.} 
        \label{fig_motivation2}
        \vspace{0em}
\end{figure*}

In the realm of computer vision, the rapid progress of convolutional neural networks (CNNs) trained on balanced datasets has led to remarkable advancements~\cite{krizhevsky2012imagenet,noh2015learning,ouyang2016deepid}. However, real-world scenarios frequently involve large-scale datasets characterized by imbalanced and long-tailed distributions~\cite{long_tail_learning1, long_tail_learning2, long_tail_learning3, long_tail_learning_survey}. In long-tail distributions, the categories with a majority of samples are termed as \textit{long} classes, while those with fewer samples are termed \textit{tail} classes. This inherent data distribution imbalance poses a significant challenge when training models for computer vision tasks. Within this context, tail classes encounter substantial under-representation during the training process, negatively impacting its recognition performance for these minority categories~\cite{long_tail_learning_survey}. Furthermore, the model tends to exhibit bias towards long classes, due to extensive training data available for these majority categories.

Moreover, in real-time applications, not all class categories are concurrently accessible; data becomes available in a continuous manner, and previous classes data might not be accessible due to privacy or storage limitations~\cite{li2017learninglwf}. Expanding our model's knowledge to encompass this continuously evolving data is of paramount importance. In the existing literature, the process of incrementally adding these classes to deep neural networks is referred to as class incremental learning (CIL)~\cite{rebuffi2017icarl}. In this context, the addition of a set of new class information into the model is termed a \textit{task}. At the end of each task in CIL, the model is evaluated on all the classes encountered so far. Typically, the initial task is trained using the cross-entropy (CE) loss and is often referred to as the base task, and gradually new classes are added at each incremental task.

Recently, Liu et al. \cite{longtail_cil} introduced long-tail distributions into the domain of class incremental learning (CIL) and coined the term \textit{``long-tail class incremental learning"}. This approach involves the model's endeavour to progressively learn new classes without succumbing to catastrophic forgetting of previously learned classes from the long-tailed data distributions at every task. Liu et al. \cite{longtail_cil} introduced a two-stage approach to address the challenges in long-tail CIL, where at each incremental task, the model learns through two stages. In the initial stage, they employed conventional incremental learning  methods like UCIR~\cite{ucir} or PODNET~\cite{douillard2020podnet}. Subsequently, in the second stage, they fixed the model parameters and trained an additional layer, the learnable weight scaling (LWS) layer, using a balanced dataset to address the issues in long-tail CIL. 

To better understand the effectiveness of this two-stage LWS framework, we conducted experiments on two long-tail distributions as proposed by Liu et al.~\cite{longtail_cil} on the CIFAR100~\cite{cifar} dataset. 
Here, we take 50 classes from the CIFAR100 dataset and partitioned it into two categories: long classes (25 classes) and tail classes (25 classes), based on the number of samples available. Upon fine-tuning the model with the LWS layer using CE loss, we observed a reduction in performance on the long classes and a concurrent improvement in performance on the tail classes across both long-tail scenarios from Figure \ref{fig_motivation1}.

Inherent under-representation of tail classes within long-tail representations often results in misaligned or inadequately defined classifier boundaries. Adjusting these boundaries with balanced data samples can adversely affect the performance of long classes. To tackle this challenge in the context of class incremental learning (CIL), we propose a novel two-stage framework, termed \textbf{GVAlign} ({\bf G}lobal {\bf V}ariance-Driven Classifier {\bf Align}ment). In this framework, during the second stage, we propose aligning all classifiers based on global variance and class prototypes, thus eliminating the need for balanced data (which compels the model to repeatedly encounter the same data for tail classes) or additional layers. This global variance, as an informative measure, effectively captures class properties and it is intuitive to align the classifiers based on this information. Importantly, incorporating this approach not only preserves performance on long classes but also enhances performance on tail classes. Figure~\ref{fig_motivation2} illustrates the classifier alignment of our proposed approach. Achieving such alignment of the classifier through global variance requires the presence of robust features and distinct class separations marked by smoother boundaries. To meet this prerequisite, we introduce the incorporation of mixup classes during the initial stage. This strategic addition contributes to the cultivation of robust feature representations, ultimately enhancing the approach's effectiveness.

Figure~\ref{fig_motivation1} demonstrates that the proposed classifier alignment strategy, coupled with robust feature learning, enhances the performance of tail classes without reducing the performance of long classes. This implies an improvement in the overall performance across all classes.
\color{black} 
The paper makes the following contributions:
\begin{itemize}
    \item We introduce a novel two-stage approach, termed \textbf{GVAlign} ({\bf G}lobal {\bf V}ariance-Driven Classifier {\bf Align}ment) encompassing robust feature learning in the first stage and classifier alignment in the second stage using global variance as a informative measure to address the issues in long-tail class incremental learning.
    \item Extensive experiments conducted on datasets CIFAR-100 and ImageNet-Subset demonstrate the effectiveness of our proposed approach over state-of-the-art methods across various long-tail CIL settings.
\end{itemize}

\section{Related works}
This section summarises the works related to incremental and long-tail learning.
\subsection{Class Incremental Learning}
Class Incremental Learning (CIL) aims to progressively acquire knowledge about new classes without relying on task-specific information. However, learning from newly annotated class data with abundant samples presents the challenge of catastrophic forgetting, where the model forgets the representations of old class data. The CIL approaches in the literature can be categorized into three groups based on their strategies to mitigate the problem of catastrophic forgetting: 1) \emph{Regularization-based} methods~\cite{zenke2017continualsi,kirkpatrick2017overcomingewc,aljundi2018memorymas,paik2020overcomingnpc} incorporate penalty-based loss terms at each incremental step on the learnable model weights according to their importance. 2) \emph{Distillation-based} Recent CIL approaches adopt distillation-based methods, using teacher-student distillation loss~\cite{hinton2015distilling} to mitigate catastrophic forgetting. In Learning without Forgetting (LwF) \cite{li2017learninglwf}, distillation loss is used alongside cross-entropy. Similarly, iCaRL \cite{rebuffi2017icarl} combines distillation loss with older-task exemplars selected through herding. Methods like BiC \cite{wu2019largebic} introduce new-class bias correction layers, and LwM \cite{dhar2019learninglwm} introduce information-preserving penalties or attention loss to counter model bias towards new classes. UCIR \cite{ucir} combines distillation loss with cosine normalization and inter-class separation constraint, while PODNET \cite{douillard2020podnet} proposes polled distillation loss to address catastrophic forgetting. Some recent works~\cite{zhu2021prototype,yu2020semanticsdc,zhu2021classdual} focus on non-exemplar-based methods without access to old class exemplars. 3) \emph{Architecture-based methods}~\cite{hung2019compactingarch1,rusu2016progressivearch2,yoon2017lifelongarch3,abati2020conditionalarch4,mallya2018piggybackarch5, mallya2018packnetarch6} These methods modify the network's width and depth at each incremental step. Network expansion is often proposed to learn new tasks, but this can be computationally intensive. An alternative approach is to select sub-networks from the entire architecture using masks \cite{abati2020conditionalarch4,mallya2018piggybackarch5,mallya2018packnetarch6}, storing the learned masks in memory. However, these methods require task-specific labels at inference time, which may not always be available in practical scenarios.

In this work, both UCIR~\cite{ucir} and PODNET~\cite{douillard2020podnet} serve as stage 1 baselines in the context of Long-Tail CIL. However, the proposed approach can also function as a module within other CIL methods.
\subsection{Long-Tail Learning}
The long-tailed learning problem has garnered extensive attention due to the prevalence of data imbalance issues in real-world scenarios. To tackle this challenge, various approaches have been explored to mitigate the disparity between the distribution of majority and minority classes. Some of the prominent techniques are: 1) \emph{Data Processing Methods}~\cite{chawla2002smote_dp_da1,chou2020remix_dp_da2,chu2020feature_dp_da3,han2005borderline_dp_da4,he2008adasyn_dp_da5} such as over-sampling to amplify tail data, under-sampling to reduce head data, and data Augmentation to extend tail data. 2) \emph{Class-level Re-weighting}~\cite{cui2019class_cw_1, hong2021disentangling_cw_2, huang2016learning_cw_3, tan2020equalization_cw_4} involves assigning different weights to classes to prioritize learning from the tail classes. Another approach, 3) \emph{Decoupling}~\cite{zhong2021improving_dc_1,kang2019decoupling_dc_2,zhang2021distribution_dc3},  also referred to as a two-stage approach, involves separating representation learning and classifier learning into distinct stages to enhance performance on tail classes.
\subsection{Long-Tail Incremental Learning}
\begin{figure*}[t]
    \centering
     
        \includegraphics[width=\linewidth]{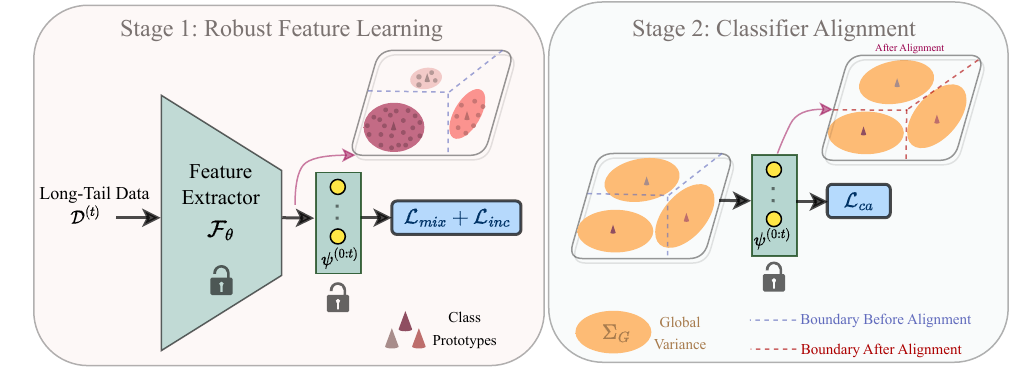}
        \caption{In stage 1, the model is trained using incremental learning approaches' loss $\mathcal{L}_{inc}$, supplemented by the mixup loss $\mathcal{L}_{mix}$, to obtain robust features. In stage 2, the feature extractor remains fixed, and only the classifiers are fine-tuned using the global variance and prototypes of the classes using classifier alignment loss $\mathcal{L}_{ca}$.}
        
    \label{main_figure_2stage}
    \vspace{-0em}
\end{figure*}
Recently, Liu et al. \cite{longtail_cil} introduced long-tail distributions into class incremental learning. They drew inspiration from the decoupling strategy's learnable weight scaling (LWS) approach \cite{kang2019decoupling_dc_2}, wherein an additional two-stage process involves training added layers using a balanced data-loader. This strategy necessitates careful design of learning approaches to effectively learn these supplementary weights \cite{kang2019decoupling_dc_2}.

In our proposed GVAlign approach, we also employ a two-stage strategy. However, distinct from the aforementioned method, we align the classifiers using prototypes and covariance without the need for a balanced dataloader or additional layers. Our strategy is more generalised due to the exploration of the feature space through sampled data points using global variance as informative measure, resulting in improved alignment of the classifiers without compromising on long classes' performance. Next, we will discuss the proposed methodology.

\section{Problem Definition and Motivation}

In this section, we begin by providing a clear explanation of the notations used in this paper. Subsequently, we delve into our two-stage training approach.
In class incremental learning, the model is sequentially trained on a total of $T$ tasks, with its data stream denoted as $\{\mathcal{D}^{(0)}, \mathcal{D}^{(1)}, ..., \mathcal{D}^{(T)}\}$, and the corresponding classes set represented by $\{\mathcal{C}^{(0)}, \mathcal{C}^{(1)}, ..., \mathcal{C}^{(T)}\}$. One assumption in CIL is there are no common classes across different tasks i.e. $\mathcal{C}^{(k)} \cap \mathcal{C}^{(l)} = \emptyset$ when $k \neq l$. At each task $t$, the model has access to data $\mathcal{D}^{(t)} = \{x_{i}, y_{i}\}_{i = 1}^{n^{(t)}}$, where $n^{(t)}$ represents the number of samples in $\mathcal{D}^{(t)}$, and $y_{i} \in \mathcal{C}^{(t)}$. In CIL, the number of samples for each class is the same and equal to $\frac{n^{(t)}}{|\mathcal{C}^{(t)}|}$; however, in the case of long-tail CIL, the data distribution of $\mathcal{D}^{(t)}$ adheres to a long-tail distribution. In both CIL and long-tail CIL, during training task $t$, alongside $\mathcal{D}^{(t)}$, the model has access to an exemplar bank $\mathcal{E}$ comprising a limited number of samples from earlier classes $\mathcal{C}^{(0:t-1)}$ and the end goal after task $t$ is to classify all the classes seen so far i.e. $\mathcal{C}^{(0:t)}$. The addition of exemplar bank $\mathcal{E}$ creates more imbalance in training data and makes long-tail CIL more challenging.


The model that learns sequentially is denoted as $\Theta = \{ \mathcal{F}_{\theta}, \psi\}$, where $\mathcal{F}_{\theta}$ represents the feature extractor with parameters $\theta$, and classifiers are represented by $\psi$. While the number of parameters in the feature extractor remains constant, the classifier layer parameters are incrementally added for each new task to accommodate novel classes. During the training task $t$, new classifiers $\psi^{(t)}$ are introduced alongside existing classifiers $\psi^{(0:t-1)}$ to classify all the classes seen so far. By using the training with data $\{\mathcal{D}^{(t)} \cup \mathcal{E}\}$, at the end of task $t$, the trained model  $\Theta^{(t)} = \{ \mathcal{F}_{\theta}, \psi^{(0:t)}\}$ is able to classify all classes from $\mathcal{C}^{(0:t)}$.

\section{Proposed GVAlign Framework }
In traditional long-tail learning, two-stage methods have shown promising results~~\cite{zhong2021improving_dc_1,kang2019decoupling_dc_2,zhang2021distribution_dc3}. These approaches decouple the feature extraction in the first stage and classifier tuning in the second stage using balanced sampling techniques~\cite{kang2019decoupling_dc_2}. However, the direct application of such methodologies to the context of long-tail CIL encounters challenges posed by catastrophic forgetting~\cite{longtail_cil}. To tackle the issues in long-tail CIL, we introduce a novel approach that entails learning robust feature representations in the first stage and refining classifier alignment in the second stage to mitigate the class imbalance problem. In the following sections, we elaborate on our stage 1 and stage 2 training procedures. Figure~\ref{main_figure_2stage} shows the overall idea of our proposed two-stage approach.
\subsection{Stage 1: Robust Feature Learning}
Inspired by~\cite{longtail_cil}, we have incorporated conventional CIL techniques, such as UCIR~\cite{ucir} and PODNET~\cite{douillard2020podnet}, into stage 1 to address class incremental learning. However, it's essential to note that our approach is not limited to these specific methods; we are adaptable to any CIL technique for stage 1. In each incremental task, the loss computed by CIL methods is denoted as $\mathcal{L}_{inc}$. As mentioned earlier, robust feature space representations are crucial for effectively tackling long-tail CIL challenges. To bolster the robustness of features at this stage, in conjunction with $\mathcal{L}_{inc}$, we propose the utilization of mixup loss~\cite{zhang2017mixup}. This implicitly accounts for the incremental stages and complements the classifier tuning stage.

Suppose we have $(x_{m}, y_{m})$ and $(x_{n}, y_{n})$ from $\mathcal{D}^{(t)}$, where $x_{m}$ and $x_{n}$ represent images and $y_{m}$ and $y_{n}$ are one-hot labels. We formulate the mixup samples and labels as follows:
\begin{equation}
\widetilde{x} = \lambda x_{m} + (1-\lambda)x_{n}
\end{equation}
\begin{equation}
\widetilde{y} = \lambda y_{m} + (1-\lambda)y_{n}
\end{equation}

Here, $\lambda$ is drawn from a Beta distribution, i.e., $\lambda \sim Beta(1,1)$. The mixup loss is then calculated as $\mathcal{L}_{mix} = \mathcal{L}_{CE}(\widetilde{x}, \widetilde{y})$, where $\mathcal{L}_{CE}(x, y) = - \sum_{k=1}^{K} y_{k} \log {(p_{x})_{k}}$ is the cross-entropy loss calculated for for $K$ classes. $p_{x} = \Theta^{(t)}(x)$ represents softmax outputs of the model. The total loss for stage 1 is 
\begin{equation}
    \mathcal{L}_{stage1} = \mathcal{L}_{inc} + \mathcal{L}_{mix}
\end{equation}

\subsection{Stage 2: Global Variance-Driven Classifier Alignment}
In the second stage, we use the class prototypes and the estimated global variance to perform the classifier alignment as described below. \\
\textbf{Construction of Proto Bank}:
Following the completion of stage 1 training, we proceed to compute a comprehensive class prototype bank denoted as $\mathcal{P}$. The objective of this prototype bank is to facilitate the alignment of classifiers in stage 2 and constructed using prototypes of all classes seen so far. Specifically, for each class $k$, the corresponding class prototype $P_{k}$ is calculated using the following equation:
\begin{align}
    P_{k} = \frac{1}{N_{k}}\sum_{\{\mathbf{x},y\} \in (\mathcal{D}^{(t)} \cup \mathcal{E})} \mathbb{I}_{(y=k)} \ \mathcal{F}_{\theta} (\mathbf{x})
\end{align}
where $N_{k}$ represents the number of samples in $k^{th}$ class, and the indicator variable $\mathbb{I}_{(y=k)}$ equals 1 if the sample belongs to the $k^{th}$ class (i.e. $y=k$). \\ \\
\textbf{Estimation of Global Variance}: 
In scenarios with long-tail data distributions, relying on tail classes for variance calculation can result in an inaccurate variance estimate that does not accurately capture the central tendencies of the class. To address this, the proposed approach uses the class with the largest number of samples during the base task ($t=0$) for global variance estimation. By aligning all classifiers based on this reliable estimate, we significantly enhance the model's capacity to generalize and discriminate across diverse class distributions. The global variance $\Sigma_{G}$ is computed as follows
\begin{align}
   \Sigma_{G} = \frac{1}{N_{G}-1} \sum_{i=1}^{N_{G}} (X_i - \bar{X})(X_i - \bar{X})^T 
\end{align}
where $X$ is the matrix that contains data points from the class with the highest number of samples and $\bar{X}$ is the mean vector of those samples. $N_{G}$ is the number of samples in that class. \\ \\
\begin{algorithm}[t]
    \caption{Proposed GVAlign Framework for Long-Tail Class Incremental Learning}
    \KwIn{ $\Theta = \{ \mathcal{F}_{\theta}, \psi\}\leftarrow$ Model\;  $\{ \mathcal{D}^{(0)}, \mathcal{D}^{(1)},..,\mathcal{D}^{(T)} \} \leftarrow $Data stream\;$e_1$$\leftarrow$ No.of epochs in stage 1\; $e_2$$\leftarrow$ No.of epochs in stage 2\; 
    $\mathcal{E} = \{ \}$ $\leftarrow$ Empty exemplar buffer}\;
    
    \For{$t \leftarrow 0$ to $T$}{
        $\mathcal{D}^{(t)} = \{ x_{i}, y_{i}  \}_{i=1}^{N_{t}}$\;
        \For{$e \leftarrow 1$ to $e_1$}{
        $\mathcal{B} = $ SampleMiniBatch$(\mathcal{D}^{(t)} \cup \mathcal{E})$\;
        $\mathcal{O}^{(0:t)} = \psi^{(0:t)}(\mathcal{F}_{\theta}(\mathcal{B}))$\;
        $\mathcal{L}_{inc} = $ IncrementalLoss$(\mathcal{B}, \mathcal{O}^{(0:t)})$\;
        $\mathcal{L}_{mix} = $ MixUpLoss$(\mathcal{B})$\;
        $\Theta \leftarrow$ UpdateParameters$(\mathcal{L}_{inc}+\mathcal{L}_{mix})$\;
        
        
    }
    $\mathcal{P} \leftarrow$ CalculatePrototypes$(\mathcal{D}^{(t)}\cup \mathcal{E})$\;

    \If{t=0}{
    $\Sigma_{G}$ = GlobalVaraince$(\mathcal{D}^{(0)})$\;
    }

    \For{$e \leftarrow 1$ to $e_2$}{
        $\mathcal{P}^\prime \leftarrow$ SampleProtoFromGlobalVar$(\mathcal{P}, \Sigma_{G})$\;
        $\mathcal{O}^{(0:t)} = \psi^{(0:t)}(\mathcal{P}^{\prime})$\;
        $\mathcal{L}_{ca} = $ ClassifierAlignLoss$(\mathcal{O}^{(0:t)})$\;
        $\psi^{(0:t)} \leftarrow$ UpdateParameters$(\mathcal{L}_{ca})$\;
        
        
    }
    $\mathcal{E} \leftarrow$ UpdateExemplars$(\mathcal{D}^{(t)})$\;
    }
    \Return{$\Theta$}\;
    \label{Algotithm}
\end{algorithm}
\begin{table*}[!h]
\centering
\begin{adjustbox}{max width=\linewidth}
\begin{tabular}{ccccc|cccc}
\hline \hline
\textit{Long-tail distribution type} $\rightarrow$&\multicolumn{4}{|c|}{\rule{-2pt}{10pt} Ordered long-tail}                                                                                                                         & \multicolumn{4}{c}{Shuffled long-tail}                                                                         \\ \hline
\multicolumn{1}{c|}{\multirow{2}{*}{\textit{Method} $\downarrow$}} & \multicolumn{2}{c}{\rule{-2pt}{10pt} CIFAR-100}                               & \multicolumn{2}{c|}{ImageNet-Subset$^{*}$}    & \multicolumn{2}{c}{CIFAR-100}                               & \multicolumn{2}{c}{ImageNet-Subset$^{*}$}    \\ \cline{2-9} 
\multicolumn{1}{c}{}                                 & \multicolumn{1}{|c}{5 tasks} & \multicolumn{1}{c}{10 tasks} & \multicolumn{1}{c}{5 tasks} & 10 tasks & \multicolumn{1}{c}{\rule{-2pt}{10pt} 5 tasks} & \multicolumn{1}{c}{10 tasks} & \multicolumn{1}{c}{5 tasks} & 10 tasks \\ \hline \hline 

\multicolumn{1}{c|}{\rule{-2pt}{10pt}  UCIR  }                                & \multicolumn{1}{c}{42.69}        & \multicolumn{1}{c}{42.15}         & \multicolumn{1}{c}{56.45 }        &    55.44      & \multicolumn{1}{c}{35.09}       & \multicolumn{1}{c}{34.59}         & \multicolumn{1}{c}{46.45}        & 45.31        \\ 

\multicolumn{1}{c|}{UCIR + LWS (ECCV 2022) }                                & \multicolumn{1}{c}{45.88}        & \multicolumn{1}{c}{45.73}         & \multicolumn{1}{c}{57.22}        &     55.41     & \multicolumn{1}{c}{39.40}       & \multicolumn{1}{c}{39.00}         & \multicolumn{1}{c}{49.42}        & \textbf{47.96}          \\ 

\multicolumn{1}{c|}{UCIR + GVAlign (Ours) }                                & \multicolumn{1}{c}{\textbf{47.13}}        & \multicolumn{1}{c}{\textbf{46.82}}         & \multicolumn{1}{c}{\textbf{58.08} }       &     \textbf{56.68 }    & \multicolumn{1}{c}{\textbf{42.80}}       & \multicolumn{1}{c}{\textbf{41.64} }        & \multicolumn{1}{c}{\textbf{50.69}}        &      47.58    \\ 

\hline 

\multicolumn{1}{c|}{\rule{-2pt}{10pt}  PODNET  }                                & \multicolumn{1}{c}{44.07}        & \multicolumn{1}{c}{43.96}         & \multicolumn{1}{c}{59.16}        &   57.74       &  \multicolumn{1}{c}{36.64}       & \multicolumn{1}{c}{34.84}         & \multicolumn{1}{c}{47.61}        & 47.85          \\ 

\multicolumn{1}{c|}{PODNET + LWS (ECCV 2022) }                                & \multicolumn{1}{c}{44.38}        & \multicolumn{1}{c}{44.35}         & \multicolumn{1}{c}{60.12}        & 59.09        & \multicolumn{1}{c}{36.37}       & \multicolumn{1}{c}{37.03}         & \multicolumn{1}{c}{49.75}        &49.51          \\ 

\multicolumn{1}{c|}{PODNET + GVAlign (Ours) }                                & \multicolumn{1}{c}{\textbf{48.41}}        & \multicolumn{1}{c}{\textbf{47.71}}         & \multicolumn{1}{c}{\textbf{61.06}  }      & \textbf{60.08 }        & \multicolumn{1}{c}{\textbf{42.72} }      & \multicolumn{1}{c}{\textbf{41.61}}         & \multicolumn{1}{c}{\textbf{52.01}  }      &\textbf{50.81 }       \\ 

\hline \hline 
\end{tabular}
\end{adjustbox}
\caption{Experimental results on long-tail class incremental learning (\textbf{$*$} signifies that we have rerun all experiments on the Imagnet-Subset 100 dataset. Comprehensive dataset details and data can be found in the GitHub repository for reproducibility).}
\vspace{-1em}
\label{long_tal_main_exp}
\end{table*}
{\bf Classifier Alignment:} 
This stage 2 training involves leveraging the computed global variance $\Sigma_{G}$ as an informative measure to explore the feature space around the prototypes $\mathcal{P}$ of all classes. This exploration aids in aligning the classifiers effectively, facilitating improved classification performance. At this stage only classifiers are tuned using pseudo-augmented samples $\mathcal{P}^\prime \sim \mathcal{N}(\mathcal{P}, \Sigma_{G})$ generated from normal distribution by employing prototypes as means and the global variance as covariance information.
The classifier alignment loss calculated during this stage as follows 
\begin{equation}
    \mathcal{L}_{ca} =  -  \sum_{q \ \in \ \mathcal{P}^{\prime} } \sum_{k=1}^{K} \hat{y}_{k} \log ({\psi^{(0:t)}(q))_{k}}
\end{equation}
where $\hat{y}$ represents the prototype one-hot label and $K$ represents the all classes seen so far.
The exemplar set $\mathcal{E}$ is updated using herding~\cite{rebuffi2017icarl} technique. which is a commonly employed technique in CIL approaches to store exemplars. The complete training procedure is summarized in Algorithm~\ref{Algotithm}.

\section{Experiments}
In this section, we discuss the datasets used, implementation details, and the results obtained in both long-tail and conventional CIL settings.
\subsection{Datasets and Evaluation Protocol} To evaluate the efficacy of our proposed framework, we conducted experiments using two benchmark datasets specifically designed for long-tail CIL~\cite{longtail_cil}: CIFAR100 and the ImageNet Subset. For a comprehensive and fair comparison, we adopted the data task splits recommended in \cite{longtail_cil}, utilizing 50 classes for the base task. Then in the 5-task configuration (T=5), we progressively introduced 10 new classes during each incremental task i.e. $(50-10- \cdot \cdot \cdot -10)$. Similarly, in the 10-task setup (T=10), we incorporated 5 new classes in each incremental task i.e. $(50-5-\cdot \cdot \cdot-5)$. Our approach followed the same long-tail distributions as proposed in \cite{longtail_cil}. \\
\textbf{CIFAR-100}:
This dataset comprises 50,000 training images and 10,000 test images, each image consisting of 32x32 pixels. These images are distributed across 100 classes.\\
\textbf{ImageNet Subset}:
The ImageNet Subset consists of 100 classes, sampled from the larger ImageNet dataset~\cite{krizhevsky2012imagenet}. All images were resized to 256x256 pixels and subsequently randomly cropped to 224x224 pixels during the training phase. We evaluated all methods on this dataset to ensure reliable and accurate evaluation.

We employ the widely recognized CIL evaluation metric, average incremental accuracy~\cite{rebuffi2017icarl, lopez2017gradientGEM}. Here, let $t$ represent the task ID, where $t\in{0,1,...,\mathcal{T}}$. We define $Acc_{0:n}^{t}$ as the model's accuracy on the test data of all tasks from $0$ to $n$ after learning task $t$, where $n \leq t$. Consequently, upon completion of task $T$, the average incremental accuracy is computed as $\frac{1}{T}\sum_{t=0}^{T} Acc_{0:t}^{t}$.
We utilized the same model architectures as in~\cite{longtail_cil} to ensure a fair comparison. Specifically, ResNet-32 was employed for CIFAR-100, while ResNet-18 was chosen for the ImageNet Subset dataset.

Our training protocol involved initiating the learning rate at 0.1 and subsequently reducing it by a factor of 10 after the $250^{th}$, $350^{th}$, and $450^{th}$ epochs, resulting in a total $500$ epochs for CIFAR-100 training. As for the ImageNet Subset, the learning rate was set to 0.1 at the start and reduced by a factor of 10 after the $30^{th}$ and $60^{th}$ epochs, resulting in a total of $90$ epochs for training. Throughout all experiments, a fixed batch size of 128 was used.
During 2-stage classifier alignment training, we tuned only classifier layers with a learning rate of 0.1 for 100 epochs. We considered a standard of 20 exemplars for each class to ensure a fair comparison with other methods. We used NVIDIA RTX A5000 24GB card to run all our experiments.

\subsection{Results on Long-Tail CIL}
First, we report the results on the long-tailed CIL task, which is the main focus of this work. 
We integrate the proposed GVAlign framework with UCIR and PODNET as in~\cite{longtail_cil} and compare with the state-of-the-art approach~\cite{longtail_cil}, which is the only work which addresses the challenging long-tailed CIL to the best of our knowledge. 
We observe from Table~\ref{long_tal_main_exp} that across both the datasets (CIFAR100 and ImageNet Subset) and different task setups (T=5 and T=10), our approach consistently achieves higher average incremental accuracy over the state-of-art long tail CIL method. 
This improvement is consistent for both ordered and shuffled long-tail distributions. Specifically, when combined with the UCIR approach on shuffled long-tail distributions, our method boosts CIFAR100 accuracy by $3.4\%$ and ImageNet Subset accuracy by $1.2\%$ in the 5-task scenario. On ordered long-tail distributions, we see a $1.25\%$ increase for CIFAR100 and a $0.86\%$ increase for ImageNet Subset in the same 5-task scenario. Notably, these gains are even more significant when PODNET serves as the baseline for CIL. With shuffled long-tail distributions, our approach achieves a remarkable $6.35\%$ improvement on CIFAR100 and a substantial $2.26\%$ increase on ImageNet Subset in the 5-task context. Similar improvements are seen in ordered long-tail scenarios, with gains of $4.03\%$ on CIFAR100 and $0.94\%$ on ImageNet Subset.

\begin{table}[t]
\centering
\begin{adjustbox}{max width=\linewidth}
\begin{tabular}{lllll}
\hline \hline
\multirow{2}{*}{\textit{Method}} & \multicolumn{2}{c}{\rule{-2pt}{10pt} CIFAR 100}          & \multicolumn{2}{c}{ImageNet-Subset$^{*}$}    \\ \cline{2-5} 
                                 & \multicolumn{1}{l}{\rule{-2pt}{10pt} 5 tasks} & 10 tasks & \multicolumn{1}{l}{5 tasks} & 10 tasks \\ \hline \hline \rule{-2pt}{10pt} 
               UCIR      & \multicolumn{1}{l}{61.15}        &    58.74      & \multicolumn{1}{l}{69.11}        &65.15       \\ 
   
    UCIR + LWS (ECCV 2022)     & \multicolumn{1}{l}{63.48}        &     60.57     & \multicolumn{1}{l}{68.83}        &66.47         \\ 
    UCIR + GVAlign (Ours)   & \multicolumn{1}{l}{\textbf{64.11}}        &     \textbf{61.23}     & \multicolumn{1}{l}{\textbf{70.05} }       &\textbf{66.60}         \\ 
 
    \hline \rule{-2pt}{10pt}
 
  PODNET      & \multicolumn{1}{l}{63.15}        &       61.16   & \multicolumn{1}{l}{67.92}        &    62.39     \\ 

    PODNET + LWS (ECCV 2022)     & \multicolumn{1}{l}{64.58}        &       62.63  & \multicolumn{1}{l}{\textbf{69.43}}        &  62.12      \\ 
    PODNET + GVAlign (Ours)     & \multicolumn{1}{l}{\textbf{65.73}}        &       \textbf{63.72}   & \multicolumn{1}{l}{68.85}        &  \textbf{62.42}        \\ 
 \hline \hline
\end{tabular}
\end{adjustbox}
\caption{Experimental results on traditional class incremental learning (\textbf{$*$} signifies that we have rerun all experiments on the Imagnet-Subset 100 dataset).}
\label{conv_cil_main_exp}
\vspace{-1em}
\end{table}
\subsection{Results on Conventional CIL}
Conventional CIL is also inherently imbalanced, since during the incremental stages, there might be very few exemplars from the earlier classes available along with large number of examples of the new classes. 
Thus, it is important to also evaluate the effectiveness of the approaches developed for long-tailed CIL setting for the conventional CIL scenario. 
Table~\ref{conv_cil_main_exp} reports the average incremental accuracy achieved by the proposed GVAlign framework in conventional CIL setups. 
We observe that the proposed approach doesn't just excel in long-tail distributions; it also improves conventional CIL. 
\captionsetup[figure]{skip=7pt}
\begin{figure}[t]
    \centering
    \begin{subfigure}{0.45\textwidth}
        \centering
        \includegraphics[width=\linewidth]{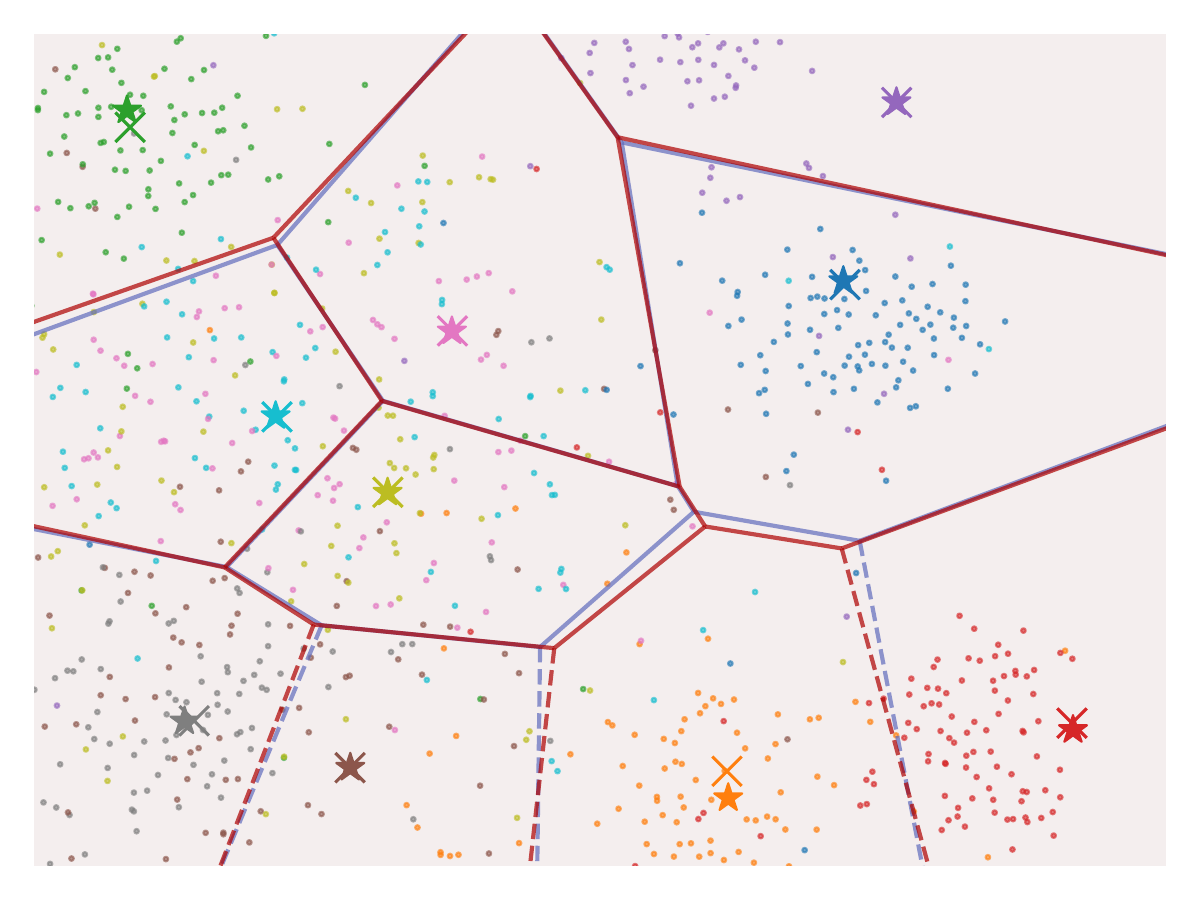}
        \caption{Scenario $T=5$: Addition of 10 new classes (before alignment $33.91\%$, after alignment $38.88\%$).}
    \end{subfigure}
    \hfill
    \begin{subfigure}{0.45\textwidth}
        \centering
        \includegraphics[width=\linewidth]{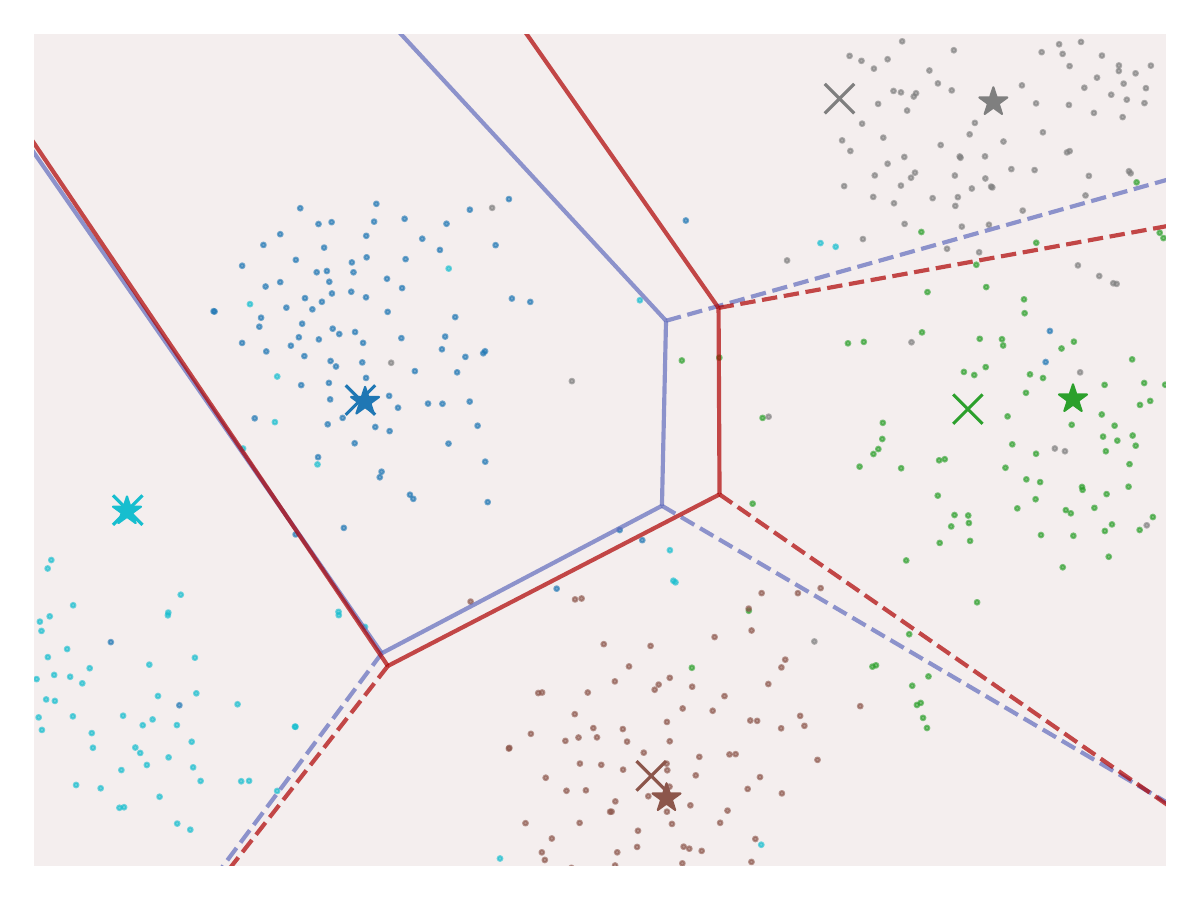}
        \caption{Scenario $T=10$: Addition of 5 new classes (before alignment $43.61\%$, after alignment $50.01\%$).}
    \end{subfigure}
    \caption{Voronoi class boundaries in the shuffled long-tail CIL scenario after task 1. The symbols `$\times$' indicate the initial classifiers and `$\star$' represents the aligned classifiers. Class boundaries before alignment are marked by \textcolor{color1_boundary1}{\rule[0.5ex]{0.5cm}{0.2mm}} and after alignment marked by \textcolor{color1_boundary2}{\rule[0.5ex]{0.5cm}{0.2mm}}.}
    \label{varonoi_plots}
    \vspace{-1em}
\end{figure}
\begin{figure*}[t]
\centering
\begin{tikzpicture}

\definecolor{color0}{HTML}{A7226E}
\definecolor{color1}{HTML}{f6aa1c}
\definecolor{color2}{HTML}{004e89}
\definecolor{color3}{HTML}{056676}
\begin{groupplot}[group style={group name =GSS CIL experiments all datasets,group size=3 by 1,horizontal sep = 1.5cm,vertical sep = 1.5cm,}, title style={anchor=north, yshift=2.5ex} ]
\nextgroupplot[
tick align=outside,title={(a) Shuffled long-tail},
x grid style={white!80!black},
x label style={at={(axis description cs:0.5,-0.2)},anchor=north},
xlabel={No.of exemplars $/$ class},
xmajorgrids,
xmin=4, xmax=21,
xtick pos=left,
xtick={5,10,15,20},
xtick style={color=white!15!black},
y grid style={white!80!black},
y label style={at={(axis description cs:-0.14,.5)}},
ylabel={Avg inc acc $(\%)$},
ymajorgrids,
ymajorticks=true,
ymin=32, ymax=44,
ytick={32,36,40,44},
ytick pos=left,
ytick style={color=white!15!black},
width = 6cm, height = 4cm,
legend to name={CommonLegend},
legend style={at={(-0.25,+1.2)},draw=none,legend columns=3, legend entries={UCIR, UCIR+LWS, UCIR+GVAlign}, legend cell align={left},/tikz/every even column/.append style={column sep=0.5cm},nodes={scale=0.55, transform shape}},mark options=solid]
\coordinate (c1) at (rel axis cs:0.80,0.5);
\addplot [thick, color0, mark=*,opacity=1, mark size=1, ]
table {
5 34.5
10 34.9
15 35.0
20 35.1
};
\addplot [thick, color1, mark=*,opacity=1, mark size=1, ]
table {
5 37.41
10 38.13
15 38.35
20 39.40
};
\addplot [thick, color2, mark=*,opacity=1, mark size=1, ]
table {
5 38.8
10 41.2
15 41.8
20 42.8
};

\nextgroupplot[
tick align=outside,title={(b) Ordered long-tail},
x grid style={white!80!black},
x label style={at={(axis description cs:0.5,-0.2)},anchor=north},
xlabel={No.of exemplars $/$ class},
xmajorgrids,
xmin=4, xmax=21,
xtick pos=left,
xtick={5,10,15,20},
xtick style={color=white!15!black},
y grid style={white!80!black},
y label style={at={(axis description cs:-0.14,.5)}},
ylabel={Avg inc acc $(\%)$},
ymajorgrids,
ymajorticks=true,
ymin=41, ymax=49,
ytick = {41,43,46,49},
ytick pos=left,
ytick style={color=white!15!black},
width = 6cm, height = 4cm,
legend to name={CommonLegend},
legend style={at={(-0.25,+1.2)},draw=none,legend columns=3, legend entries={UCIR, UCIR+LWS, UCIR+GVAlign}, legend cell align={left},/tikz/every even column/.append style={column sep=0.5cm},nodes={scale=0.75, transform shape}},mark options=solid]
\coordinate (c1) at (rel axis cs:0.80,0.5);
\addplot [thick, color0, mark=*,opacity=1, mark size=1, ]
table {
5 42.0
10 42.20
15 42.5
20 43.28
};
\addplot [thick, color1, mark=*,opacity=1, mark size=1, ]
table {
5 43.6
10 45.13
15 45.2
20 45.88
};
\addplot [thick, color2, mark=*,opacity=1, mark size=1, ]
table {
5 45.7
10 47.01
15 47.4
20 48.0
};

\nextgroupplot[
tick align=outside,title={(c) Conventional},
x grid style={white!80!black},
x label style={at={(axis description cs:0.5,-0.2)},anchor=north},
xlabel={No.of exemplars $/$ class},
xmajorgrids,
xmin=4, xmax=21,
xtick pos=left,
xtick={5,10,15,20},
xtick style={color=white!15!black},
y grid style={white!80!black},
y label style={at={(axis description cs:-0.14,.5)}},
ylabel={Avg inc acc $(\%)$},
ymajorgrids,
ymajorticks=true,
ymin=58, ymax=66,
ytick={58,60,63,66},
ytick pos=left,
ytick style={color=white!15!black},
width = 6cm, height = 4cm,
legend to name={CommonLegend},
legend style={at={(-0.25,+1.2)},draw=black,legend columns=3, legend entries={UCIR, UCIR + LWS, UCIR + GVAlign (Ours)}, legend cell align={left},/tikz/every even column/.append style={column sep=0.5cm},nodes={scale=0.85, transform shape}},mark options=solid]
\coordinate (c1) at (rel axis cs:0.80,0.5);
\addplot [thick, color0, mark=*,opacity=1, mark size=1, ]
table {
5 59.2
10 60.
15 60.5
20 61.15
};
\addplot [thick, color1, mark=*,opacity=1, mark size=1, ]
table {
5 59.71
10 61.88
15 62.2
20 63.48
};
\addplot [thick, color2, mark=*,opacity=1, mark size=1, ]
table {
5 59.6
10 62.26
15 63
20 64.11
};
\end{groupplot}

\node at (8.38,3.3)
{\pgfplotslegendfromname{CommonLegend}};

\end{tikzpicture}

\caption{Illustrates how our proposed approach consistently improves with an increased number of exemplars, benefiting from precise prototype positioning as the number of exemplars increases.}
\label{ablation_graphs}
\vspace{0em}
\end{figure*}
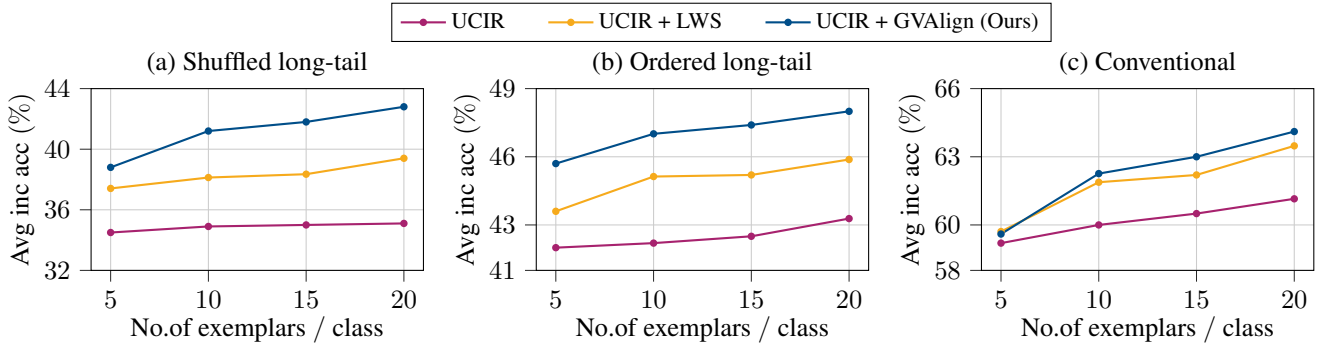
\section{Analysis $\&$ Ablation Studies}
In this section, we delve into the analysis of our proposed approach. Firstly, we analyze the alignment of classifiers using Voronoi plots~\cite{aurenhammer1991voronoi}. Next, we examine the impact of the number of exemplars in the context of long-tail CIL. Notably, our approach consistently outperforms state-of-the-art techniques in various long-tail CIL scenarios, irrespective of the number of exemplars utilized. Subsequently, we explore the benefits of our approach in a conventional setting where all new classes contain equal samples. We observe that due to robust learning and feature space exploration, our method enhances the separation between semantically similar classes. This improvement alleviates potential confusion between these classes, ultimately leading to enhanced performance.
\subsection{Analyzing Classifiers}
Our approach's key contribution lies in effectively aligning classifiers during incremental learning for long-tail data distributions. To highlight the significance of this alignment, we utilize Voronoi plots to illustrate the learning of new classes during task 1 in the context of shuffled long-tail CIL. Voronoi plots visually illustrate feature space regions assigned to different classes, providing insights into classifier behavior. We present visualizations for two different settings: $T=5$, where 10 new classes are introduced, and $T=10$, where 5 new classes are added in Figure~\ref{varonoi_plots}. The alignment of classifiers results in a tangible enhancement in accuracy for the newly introduced classes. Specifically, in the T=5 scenario, the classification accuracy on task1 improves from $33.91\%$ to $38.88\%$. Similarly, in the T=10 scenario, the accuracy rises from $43.61\%$ to $50.01\%$.

\subsection{Effect of Number of Exemplars}
To understand the effect of the number of exemplars, we conduct extensive experiments with exemplar counts of $\{ 5,10,15,20 \}$, as depicted in Figure~\ref{ablation_graphs}. Across both shuffled and ordered long-tail scenarios, our approach consistently outperforms existing long-tail CIL methods. Notably, it also exhibits strong performance in conventional CIL settings. As the number of exemplars increases, our approach's ability is even more pronounced due to the precise position of prototypes in the representation space.
\begin{figure}[t]
\centering
\begin{tikzpicture}
\definecolor{color0}{HTML}{E8DED2}
\definecolor{color1}{HTML}{A3D2CA}
\definecolor{color2}{HTML}{5EAAA8}
\definecolor{color3}{HTML}{056676}
\begin{groupplot}[group style={group size=1 by 1,  vertical sep=10ex}, title style={anchor=north, yshift=2ex}, width=5cm]
\nextgroupplot[
axis line style={white!80!black},
tick align=outside,
x grid style={white!80!black},
xmajorticks=true,
xmin=-0.4, xmax=1.4,
xtick style={color=white!15!black},
xtick={0,1},
xticklabels={Clear semantic group, Confusing semantic group},
xticklabel style={
        anchor=north, 
        yshift=0.5pt, 
        align=center,
    },
x label style={at={(axis description cs:-0.8,0.4)},anchor=north},
xtick pos=left,
xmajorgrids,
y grid style={white!80!black},
y label style={at={(axis description cs:-0.10,.5)}},
ylabel={Accuracy (\%)},
ymajorticks=true,
ymajorgrids,
ymin=60, ymax=90,
ytick style={color=white!15!black},
ytick={60,70,80,90},
ytick pos=left,
width = 8cm, height = 4.8cm,
legend style={at={(0.48,+1.1)},  legend cell align={left},/tikz/every even column/.append style={column sep=0.1cm},
anchor=south,legend columns=3},
]
\draw[draw=white,fill=color0,opacity=1] (axis cs:-0.3,0) rectangle (axis cs:-0.1,86.92);
\addlegendimage{line width=1.5mm,color=color0}
\addlegendentry{CE}
\draw[draw=white,fill=color0,opacity=1] (axis cs:0.7,0) rectangle (axis cs:0.9,64.56);
\draw[draw=white,fill=color1,opacity=1] (axis cs:-0.1,0) rectangle (axis cs:0.1,86.7);
\addlegendimage{line width=1.5mm,color=color1}
\addlegendentry{CE + LWS }
\draw[draw=white,fill=color1,opacity=1] (axis cs:0.9,0) rectangle (axis cs:1.1,65);
\draw[draw=white,fill=color3,opacity=1] (axis cs:0.1,0) rectangle (axis cs:0.3,88.59);
\addlegendimage{line width=1.5mm,color=color3}
\addlegendentry{CE + GVAlign (Ours)}
\draw[draw=white,fill=color3,opacity=1] (axis cs:1.1,0) rectangle (axis cs:1.3,68.76);

\end{groupplot}

\end{tikzpicture}

\vspace{-1em}
\caption{Shows the average accuracy of clear and confusing semantic groups.}
\vspace{-1em}
\label{semantic_sim}
\end{figure}
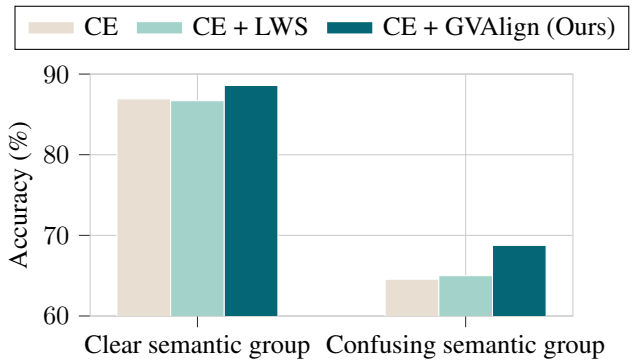
\subsection{Separating Semantic Similar Classes}
\label{Separating Semantic Similar Classes}
The difference in performance among different training classes depends not only upon the characteristics of the training data (like number of examples per class), but also on the classes themselves. 
For example, few classes maybe semantically very close~\cite{deselaers2011visual_semantic}and thus inherently confusing, which can often lead to reduced performance~\cite{chen2020simple_semantic, wang2014learning_semantic}, even with the same number of training data per class. 
The proposed framework, though developed primarily for long-tailed setting, can also seamlessly account for these other challenges, since it tries to push the classifiers apart in the second stage.
To verify this, we divide the total 50 base classes into two groups: (i) clear semantic group, where 25 classes are well-separated from the rest, and (ii) confusing semantic group, where the performance of these 25 classes is adversely affected due to confusion with other classes. This grouping is based on the sorting order of individual class performances. Figure~\ref{semantic_sim} illustrates the average accuracy of these two groups. We observe that our proposed approach significantly improves performance, particularly for the confusing semantic group, justifying its effectiveness.

\subsection{Ablation on Proposed Losses}
Table~\ref{Ablation} presents an analysis of the individual components of our proposed methodology. Clearly, the incorporation of losses at different stages contributes to the improvement of feature representations and the alignment of classifiers, leading to an enhancement in overall performance.

\begin{table}[!h]
\centering
\begin{adjustbox}{max width=\linewidth}
\begin{tabular}{ccccccc}
\hline \hline
\multicolumn{3}{c}{\rule{-2pt}{10pt}\textit{Distribution type $\rightarrow$}} & \multicolumn{2}{c}{\rule{-2pt}{10pt} Ordered long-tail}          & \multicolumn{2}{c}{Shuffled long-tail}    \\ \cline{1-7} 
           UCIR & $\mathcal{L}_{mix}$ & $\mathcal{L}_{ca}$                     & \multicolumn{1}{c}{\rule{-2pt}{10pt} 5 tasks} & 10 tasks & \multicolumn{1}{c}{5 tasks} & 10 tasks \\ \hline \hline \rule{-2pt}{10pt} 
                          
               \cmark   & \xmark & \xmark &  \multicolumn{1}{c}{42.69}        &    42.15      & \multicolumn{1}{c}{35.09}        & 34.59          \\ 
   
    \cmark & \cmark   & \xmark &   \multicolumn{1}{c}{45.31}        &     44.84     & \multicolumn{1}{c}{39.11}        &38.55          \\ 
    \cmark & \cmark  &\cmark &  \multicolumn{1}{c}{\textbf{47.13}}        &     \textbf{46.82}     & \multicolumn{1}{c}{\textbf{42.80} }       &\textbf{41.64}         \\
 \hline \hline
\end{tabular}
\end{adjustbox}
\caption{Presents an ablation study showcasing the impact of introducing different losses in our proposed methodology.}
\label{Ablation}
\vspace{-1em}
\end{table}

\section*{Conclusion}
In conclusion, this paper presents a significant advancement in addressing the challenges of long-tail class incremental learning through a novel two-stage framework. Our approach excels in both learning new classes progressively and mitigating catastrophic forgetting in the presence of imbalanced data distributions. By incorporating robust feature learning in the first stage and harnessing the power of global variance as an informative measure in the second stage, we achieve effective classifier alignment without resorting to data balancing or additional layer tuning. Extensive experimental validation on various datasets corroborates the superiority of our approach compared to SOTA methods in various long-tail class incremental learning scenarios.\\
\textbf{Acknowledgements:}
This work is partly supported through a research grant from SERB, Department of Science and Technology, Govt. of India (SPF/2021/000118).


{\small
\bibliographystyle{ieee_fullname}
\bibliography{egbib}
}

\end{document}


\title{Robust Feature Learning and Global Variance-Driven Classifier Alignment for Long-Tail Class Incremental Learning \\ (Supplementary Material)}  

\maketitle

\section{Illustration of Long-Tail CIL data distributions:}

\begin{figure*}[h]
\centering

\caption{CIFAR100 data distributions for different scenarios in class incremental learning for T=5 setting.}

\end{figure*}


\begin{figure*}
\centering
%
\caption{CIFAR100 data distributions for different scenarios in class incremental learning for T=10 setting.}

\end{figure*}